\documentclass[conference]{IEEEtran}
\IEEEoverridecommandlockouts
\usepackage{cite}
\usepackage{amsmath,amssymb,amsfonts}
\usepackage{algorithmic}
\usepackage{graphicx}
\usepackage{textcomp}
\usepackage{xcolor}

\usepackage{hyperref}
\usepackage{orcidlink}
\hypersetup{hidelinks=true}
\usepackage{textcomp}
\usepackage[numbers, sort&compress]{natbib}

\def\BibTeX{{\rm B\kern-.05em{\sc i\kern-.025em b}\kern-.08em
    T\kern-.1667em\lower.7ex\hbox{E}\kern-.125emX}}
\begin{document}

\title{ProtoSolo: Interpretable Image Classification via Single-Prototype Activation}

\author{
	\textbf{Yitao Peng\textsuperscript{1}},
	\textbf{Lianghua He\textsuperscript{1}},
	\textbf{Hongzhou Chen\textsuperscript{1}}\\
	\textsuperscript{1}School of Computer Science and Technology, Tongji University, Shanghai 201804, China\\
	Email: pyt@tongji.edu.cn, helianghua@tongji.edu.cn, chenhongzhou@tongji.edu.cn
}

\maketitle

\begin{abstract}
	Although interpretable prototype networks have improved the transparency of deep learning image classification, the need for multiple prototypes in collaborative decision-making increases cognitive complexity and hinders user understanding. To solve this problem, this paper proposes a novel interpretable deep architecture for image classification, called ProtoSolo. Unlike existing prototypical networks, ProtoSolo requires activation of only a single prototype to complete the classification. This design significantly simplifies interpretation, as the explanation for each class requires displaying only the prototype with the highest similarity score and its corresponding feature map. Additionally, the traditional full-channel feature vector is replaced with a feature map for similarity comparison and prototype learning, enabling the use of richer global information within a single-prototype activation decision. A non-projection prototype learning strategy is also introduced to preserve the association between the prototype and image patch while avoiding abrupt structural changes in the network caused by projection, which can affect classification performance. Experiments on the CUB-200-2011 and Stanford Cars datasets demonstrate that ProtoSolo matches state-of-the-art interpretable methods in classification accuracy while achieving the lowest cognitive complexity. The code is available at \href{https://github.com/pyt19/ProtoSolo}{https://github.com/pyt19/ProtoSolo}.
\end{abstract}

\section{Introduction} \label{Section_1}

Convolutional neural networks (CNNs) \cite{xu2024sctnet,liu2024unbiased} excel in computer vision tasks. However, their black-box nature impedes model interpretability \cite{peng2024decoupling}. This lack of explainability is particularly critical in high-stakes decision-making scenarios, such as medical diagnostics and autonomous driving, where understanding the decision-making process of the model is crucial for ensuring reliability, fairness, and user trust. To address this, prototype-based interpretable networks have emerged as a prominent research direction \cite{wang2023learning}. These networks develop inherently interpretable architectures by learning specific prototypes for each category and performing classification through a weighted summation of similarities between input features and prototypes. Nonetheless, current methods struggle to identify the primary prototype that drives decision-making when providing local explanations for a single category, as they must account for the combined effects of all prototypes within the category, complicating the explanation process.

\begin{figure}[!t]
	\centerline{\includegraphics[width=\columnwidth]{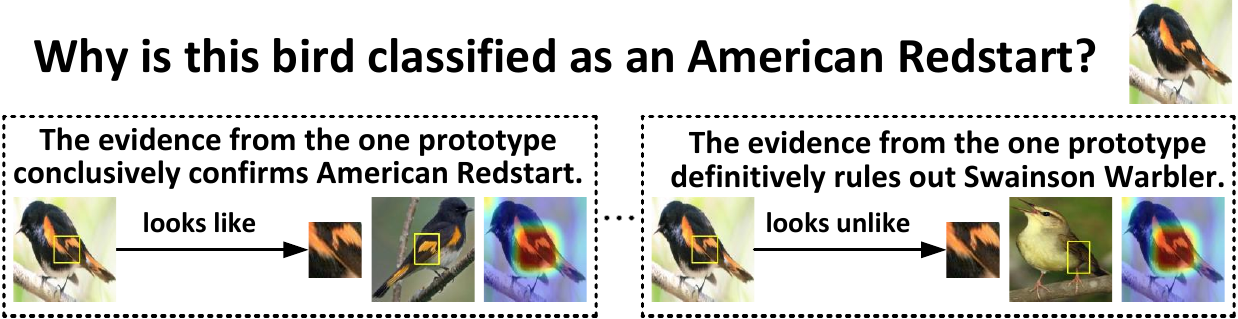}}
	\caption{Example of the American Redstart classification. ProtoSolo imitates human decision making and classifies based on the visual similarity between the key area of the input image and each class's most similar prototype.}
	\label{fig_1}
\end{figure}

Inspired by human discrimination mechanisms, where object categories are typically defined by a single distinguishing feature, this study introduces a novel interpretable architecture, ProtoSolo. Unlike traditional prototype networks that use a multi-prototype collaborative decision-making approach, ProtoSolo incorporates a maximum operation between prototype layer and fully connected layer, compelling the network to classify based solely on the prototype with the highest similarity within each category. This design enables local interpretations to concentrate on a single key prototype that governs the decision and its associated input features, thereby substantially reducing cognitive complexity. As depicted in Figure \ref{fig_1}, the input image is classified into the American Redstart category because it closely matches the most similar prototype of this category while differing from that of the Swainson Warbler category. The primary innovation of this mechanism is determining the classification probability of each category solely based on the similarity between the input features and the most similar prototype for that category, thereby establishing an interpretable decision-making framework centered on a single prototype.

\begin{figure}[!t]
	\centerline{\includegraphics[width=\columnwidth]{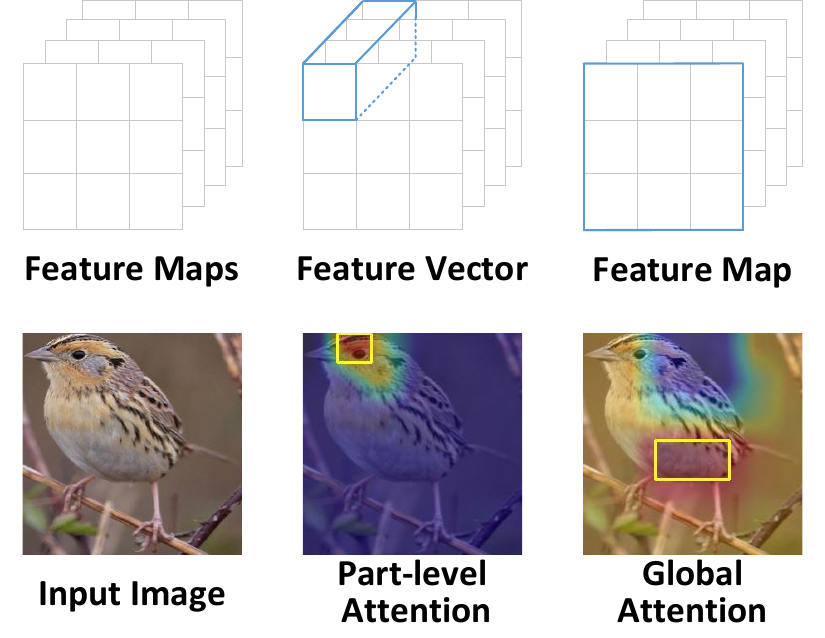}}
	\caption{Input image encoded by the CNN to obtain the feature maps from which the full-channel feature vector and feature map can be extracted. In traditional prototype networks, visual interpretation of the feature vector provides part-level attention that reflects the local information of the input image. In ProtoSolo, visual interpretation of the feature map provides global attention that reflects the global information of the input image.}
	\label{fig_2}
\end{figure}

Conventional prototype networks \cite{chen2019looks} make decisions by comparing similarities between local features and prototypes. However, these local features and prototypes provide limited information owing to the lack of global context, potentially compromising classification accuracy. To maintain the inherent interpretable reasoning structure of a prototypical network, this study introduces a feature-map-based comparison method. The approach replaces the fundamental units of similarity comparison and prototype learning from the full-channel feature vector with a feature map. The global attention within the feature map captures richer contextual information compared to the part-level attention associated with the previous full-channel feature vector (Figure \ref{fig_2}). Within the single-prototype decision framework, this method enhances classification accuracy by increasing the informational content of similarity comparison objects while preserving the interpretability of the decision.

Conventional prototype networks \cite{wang2021interpretable,ayoobi2025protoargnet} link prototypes to specific key features of training images through prototype projection operations to improve model interpretability. Nonetheless, the projection process induces significant and abrupt alterations within the network weights, resulting in diminished classification accuracy. Our experiments demonstrated that the existing ProtoPNet and our proposed ProtoSolo, when directly training the fully connected (FC) layer after training the convolutional and prototype layers while omitting the projection, led to prototypes that remained highly similar to the key features of the training images. This suggests that prototypes are strongly associated with analogous features without reliance on projection operations. Therefore, we propose non-projection prototype learning, which aims to preserve the visual interpretability of prototypes and their associated similarity features while mitigating performance degradation caused by projection. The primary contributions of this study are as follows:
\begin{itemize}
	\item We introduce ProtoSolo, an interpretable network based on a single-prototype activation mechanism. It employs only the prototype with the highest similarity activation in each category to make decisions, significantly reducing the cognitive complexity of model interpretation.
	\item We propose a feature-map-based comparison that transforms the similarity computation and prototype learning from a feature vector to a feature map containing richer information, thereby enhancing the accuracy of a single-prototype activation network.
	\item We present non-projection prototype learning, validated in both ProtoPNet and ProtoSolo, wherein prototypes continue to strongly associate with key areas of the image even after excluding the projection operation. This approach preserves the visual explanatory capability of the prototypes while preventing performance impairment resulting from substantial modifications to the network weights due to projection.
\end{itemize}

\section{Related Work} \label{Section_2}

\subsection{Post-hoc Interpretability Methods} \label{Section_21}

Post-hoc interpretability research \cite{di2025ante} focuses on translating trained model decisions into human-understandable visual representations. Gradient-based method \cite{sundararajan2017axiomatic} attributes predictions into inputs via gradient integration, enabling model debugging and insights without altering the network. DeepLIFT \cite{shrikumar2017learning} introduces a reference baseline, computes feature contributions to address gradient saturation, and offers directional information. Perturbation-based methods, such as RISE \cite{petsiuk2021black} and G-LIME \cite{li2023g}, treat the model as a black box, identifying key sensitive regions by systematically masking input areas and observing changes in predictions. The class activation map technique ShapCAM \cite{zheng2022shap} assesses input pixel importance by calculating their Shapley values independently of model gradients. Collectively, these methods aim to clarify the decision-making processes of complex models and enhance their transparency and credibility.

However, these explanation tools are solely a posteriori methods that analyze only the features influencing the final prediction of the model, without revealing its internal reasoning logic. This inherent limitation has spurred the pursuit of intrinsically interpretable model representations.

\begin{figure*}[t]
	\centering
	{\includegraphics[width=1.0\linewidth]{{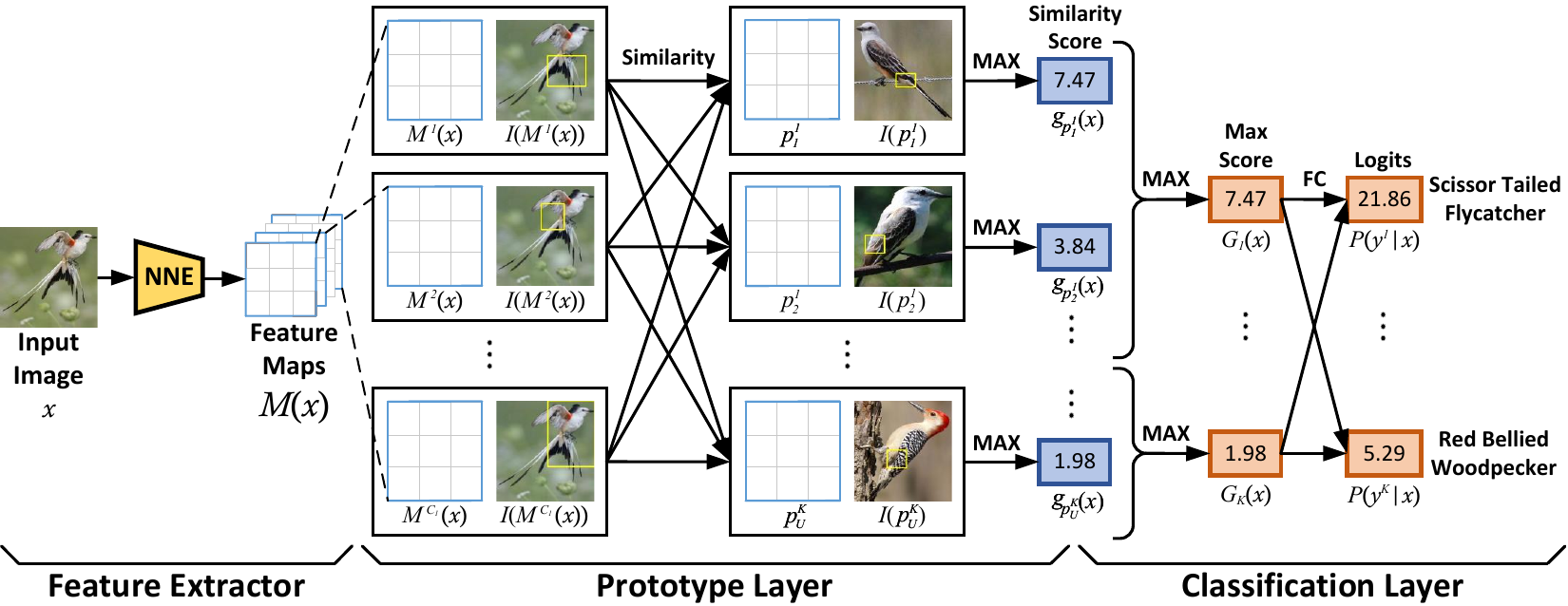}}%
	}
	\caption{Architecture of the proposed ProtoSolo method for interpretable image classification.}
	\label{fig_3}
\end{figure*}

\subsection{Interpretable Models} \label{Section_22}

Interpretable models improve transparency by incorporating inherently understandable reasoning structures within networks. Among these, prototype-based neural networks are predominant. ProtoPNet \cite{chen2019looks} uniquely learns discriminative prototypes for each category as the basis for decision-making. Subsequent advancements optimized various aspects, with TesNet \cite{wang2021interpretable} constructing a category-aware orthogonal basis concept space on the Grassmann manifold to achieve explainable classification. ProtoTree \cite{nauta2021neural} integrates prototype learning into a decision-tree framework. ProtoPShare \cite{rymarczyk2021protopshare} and ProtoPool \cite{rymarczyk2022interpretable} reduce prototype redundancy through data-dependent pruning strategies and cross-category sharing, respectively. Proto2Proto \cite{keswani2022proto2proto} introduces a prototype interpretability transfer framework utilizing knowledge distillation. Deformable ProtoPNet (D-ProtoPNet) \cite{donnelly2022deformable} incorporates prototypes with flexible spatial positions to overcome rigid limitations. PIP-Net \cite{nauta2023pip} employs self-supervised learning to align human visual prototype parts and adopts sparse scoring. ST-ProtoPNet \cite{wang2023learning} leverages SVM \cite{hearst1998support} theory to collaboratively optimize prototypes both near and distant from the decision boundary. Additionally, ProtoArgNet \cite{ayoobi2025protoargnet} aggregates prototypes into ``super prototypes'' to generate more cognitively consistent sparse evidence.

However, existing prototype-based methods have a key limitation: explaining a single category requires aggregating contributions from multiple prototypes, complicating the identification of core decision drivers. Although enforcing one prototype per class can simplify explanations, it degrades performance due to insufficient representational capacity when using traditional feature vectors. To address this, we propose ProtoSolo, which introduces (1) a single-prototype activation mechanism per class for classification, ensuring explanation clarity by design, and (2) feature-map-based comparison, wherein the feature vector is replaced by a feature map containing richer information as the comparison unit. This dual innovation facilitates highly interpretable decisions without compromising discriminative power.

\section{ProtoSolo} \label{Section_3}

ProtoSolo comprises three core modules (Figure \ref{fig_3}): the feature extractor encodes the input image into feature maps; similarity scores with the prototypes are calculated using these feature maps; and the classification layer makes decisions based on the highest similarity score. The reasoning and training processes of ProtoSolo are comprehensively detailed in the following sections.

\subsection{Feature Extractor} \label{Section_31}

The feature extractor of ProtoSolo adopts the convolutional layer from ProtoPNet \cite{chen2019looks} and comprises a neural network encoder (NNE). The NNE utilizes a sequential architecture: the backbone network $f_{b}$ (built from the CNN encoder) is followed by a shaping network $f_{s}$, which resizes feature maps. The NNE is represented by $M(\cdot)$, where $M(\cdot) = f_{s}(f_{b}(\cdot))$. An input image $x \in \mathbb{R}^{H \times W \times C}$ is processed by the NNE, producing the output feature maps $M(x) \in \mathbb{R}^{H_{1} \times W_{1} \times C_{1}}$. The matrix representation of feature maps $M(x)$ is as follows:
\begin{equation} \label{Eq_1}
M(x) = [[m_{(h,w)}^{1}(x)]_{H_1 \times W_1}, ..., [m_{(h,w)}^{C_{1}}(x)]_{H_1 \times W_1}],
\end{equation}
where $m_{(h,w)}^{c}(x)$ represents the value at spatial position $(h,w)$ and channel $c$ in the feature maps $M(x)$. $M(x)$ can be decomposed into $C_{1}$ independent two-dimensional feature maps $\{M^{c}(x)\}_{c=1}^{C_{1}}$ along the channel dimension, where $M^{c}(x)$ is the $c$-th feature map of $M(x)$.

\subsection{Feature-map-based Comparison} \label{Section_32}

To preserve the interpretability of the prototype network's reasoning process based on similarity comparisons, ProtoSolo employs the feature map set $\{M^{c}(x)\}_{c=1}^{C_{1}}$ as both the basis for similarity comparisons and the learning objects within the prototype layer. The feature map is expressed as follows:
\begin{equation} \label{Eq_2}
M^{c}(x) = [m_{(h,w)}^{c}(x)]_{H_1 \times W_1},
\end{equation}
where $M^{c}(x)$ denotes the feature map corresponding to channel index $c$ of $M(x)$. $M^{c}(x)$ captures more extensive global spatial contexts compared to traditional full-channel feature vector $M_{(h, w)}(x) = [m_{(h,w)}^{1}(x), \ldots, m_{(h,w)}^{C_{1}}(x)]$. This similarity comparison based on the feature map set $\{M^{c}(x)\}_{c=1}^{C_{1}}$ establishes a more robust foundation for the subsequent classification decision through single-prototype activation, thereby enhancing classification accuracy.

The classification task involves $K$ distinct categories. Within the ProtoSolo prototype layer, each category $k$ ($k \in \{1, 2, \dots, K\}$) is assigned $U$ prototypes, resulting in a total of $U \times K$ prototypes, denoted as $\{\{p^{k}_{u}\}_{u=1}^{U}\}_{k=1}^{K}$. Each prototype $p^{k}_{u} \in \mathbb{R}^{H_{1} \times W_{1} \times 1}$ represents the characteristic features of images belonging to category $k$ in the training dataset. The similarity function is defined as $s(\phi, \varphi) = ln(\frac{||\phi-\varphi||^{2}_{2}+1}{||\phi-\varphi||^{2}_{2}+\epsilon})$, where $\phi$ and $\varphi$ are input vectors, $\epsilon$ is a small positive constant to avoid division by zero, and $||\cdot||_{2}$ denotes the L2 norm. The function $s(\phi, \varphi)$ measures the similarity between $\phi$ and $\varphi$, with higher values indicating greater similarity. In the prototype layer, the similarity between the $u$-th prototype $p^{k}_{u}$ of category $k$ and the set of feature map $\{M^{c}(x)\}_{c=1}^{C_{1}}$ derived from the input image $x$ is calculated. The maximum similarity value is selected as the similarity score $g_{p^{k}_{u}}(x)$, representing the highest similarity between $p^{k}_{u}$ and the features of $x$. The formula is as follows:
\begin{equation} \label{Eq_3}
g_{p^{k}_{u}}(x) = \mathop{\max}\limits_{1 \leq c \leq C_{1}}s(M^{c}(x), p^{k}_{u}).
\end{equation}

After these computations, we obtain the similarity score set $\{\{g_{p^{k}_{u}}(x)\}_{u=1}^{U}\}_{k=1}^{K}$ between the input image $x$ and all category prototypes $\{\{p^{k}_{u}\}_{u=1}^{U}\}_{k=1}^{K}$, thereby determining the classification decision.

\subsection{Single-prototype Activation Classification} \label{Section_33}

To reduce the cognitive complexity of explanations, avoid decision-dependency conflicts, and identify the predominant prototype, we utilized only the prototype with the highest similarity within each category for classification. This strategy simplifies the local explanation of each prototype and its corresponding feature map. Consequently, the classification layer computes the maximum similarity score. The maximum score $G_{k}(x)$ is defined as the highest similarity score among $\{ g_{p^{k}_{u}}(x) \}_{u=1}^{U}$ generated by the input image $x$ and the prototype set $\{p^{k}_{u}\}_{u=1}^{U}$ for category $k$, as follows:
\begin{equation} \label{Eq_4}
G_{k}(x) = \mathop{\max}\limits_{1 \leq u \leq U}g_{p^{k}_{u}}(x).
\end{equation}

ProtoSolo performs classification using a fully connected (FC) layer based on the maximum score set $\{ G_{k}(x) \}_{k=1}^{K}$ (Figure \ref{fig_3}). Each category is determined solely by its highest prototype similarity score, thereby providing a concise explanation for the classification. The original logit output formula is expressed as follows:
\begin{equation} \label{Eq_5}
P(y^{t}|x) = \sum_{k=1}^{K}w^{(t,k)}G_{k}(x),
\end{equation}
where $P(y^{t}|x)$ denotes the logit predicting that input $x$ belongs to category $t$, and the weight factor $w^{(t,k)}$ indicates the contribution of the maximum score $G_{k}(x)$ from category $k$ to the prediction of category $t$.

\subsection{ProtoSolo Training} \label{Section_34}

ProtoSolo is trained on the dataset $\{ (x_{j}, y_{j}) \}_{j=1}^{n_{j}}$, where $x_{j}$ represents an image and $y_{j}$ its corresponding label. The association between training image features and prototypes were strengthened by constraining each training image feature to be near at least one prototype within its category. The clustering cost loss is defined as follows:
\begin{equation} \label{Eq_6}
L_{clst} = \frac{1}{n_{j}}\sum_{j=1}^{n_{j}} \ \mathop{min}\limits_{1 \leq u \leq U}\ \ \mathop{min}\limits_{1 \leq c \leq C_{1}}\ ||M^{c}(x_{j})-p^{y_{j}}_{u}||^{2}_{2}.
\end{equation}

To enhance the separation between prototypes and heterogeneous features, and to repel potential features from heterogeneous prototypes, the separation cost loss is defined as follows:
\begin{equation} \label{Eq_7}
L_{sep} = - \frac{1}{n_{j}}\sum_{j=1}^{n_{j}} \ \mathop{min}\limits_{\substack{1 \leq u \leq U \\ r \neq j}}\ \ \mathop{min}\limits_{1 \leq c \leq C_{1}}\ ||M^{c}(x_{j})-p^{y_{r}}_{u}||^{2}_{2}.
\end{equation}

To ensure that the prediction logit for category $t$ is derived solely from the similarity score generated by category $t$'s prototype, the weight factor loss is defined as follows:
\begin{equation} \label{Eq_8}
L_{w} = \sum_{t=1}^{K}\sum_{k \neq t}|w^{(t,k)}|.
\end{equation}

The model is trained using cross-entropy loss to ensure accurate image classification. The formula is presented as follows:
\begin{equation} \label{Eq_9}
L_{crs} = \frac{1}{n_{j}}\sum_{j=1}^{n_{j}}CrsEnt(P(x_{j}), y_{j}),
\end{equation}
where $P(x_{j})$ denotes the logits for each category corresponding to the ProtoSolo classification input $x_{j}$, and $\text{CrsEnt}(\cdot, \cdot)$ represents the cross-entropy loss function. The total loss is defined as the weighted sum of all aforementioned losses:
\begin{equation} \label{Eq_10}
L_{total} = L_{crs} + \lambda_{1}L_{clst} + \lambda_{2}L_{sep} + \lambda_{3}L_{w},
\end{equation}
where the balancing factors $\{\lambda_{i}\}_{i=1}^{3}$ act as hyperparameters, and ProtoSolo is trained by optimizing the total loss $L_{total}$.

\subsection{Non-projection Prototype Learning} \label{Section_35}

We omit the prototype projection step used in traditional methods because it introduces abrupt structure changes that impair performance. Instead, we find that prototypes trained with our loss function in equation (\ref{Eq_10}) maintain robust feature associations. Consequently, we eliminate the projection operation. This modification not only mitigates abrupt changes within the network, thereby enhancing classification accuracy, but also preserves the model's interpretability. Consider the input image $x_{\delta}$ classified by ProtoSolo into category $k_{\delta}$; the bilinear upsampling function, denoted as $B(\cdot)$, upsamples the tensor from $\mathbb{R}^{H_{1} \times W_{1} \times 1}$ to $\mathbb{R}^{H \times W \times 1}$. We propose the following method to elucidate the feature maps and prototypes involved in model classification.

Let $I(M^{c}(x_{\delta}))$ represent the visual explanation of the feature map $M^{c}(x_{\delta})$. $M^{c}(x_{\delta})$ denotes the activation intensity distribution of a specific pattern detected by a convolution kernel at a spatial location within the image $x_{\delta}$. The feature map $M^{c}(x_{\delta})$ is upsampled to match the size of $x_{\delta}$, and the significant activation region is identified using the activation value percentile threshold $\kappa$. Subsequently, the region box is superimposed onto the original image $x_{\delta}$. The corresponding formula is presented as follows:
\begin{equation} \label{Eq_11}
I(M^{c}(x_{\delta})) = B(M^{c}(x_{\delta})) \odot x_{\delta},
\end{equation}
where $\odot$ is defined as follows: based on the percentile threshold $\kappa$ of the activation map $B(M^{c}(x_{\delta}))$, the key region is identified and superimposed onto the image $x_{\delta}$.

The visual explanation $I(p^{k}_{u})$ of prototype $p^{k}_{u}$ is obtained by identifying the feature map in the training set $\{x_{j}\}^{n_{j}}_{j=1}$ that is most similar to $p^{k}_{u}$ and using the explanation of this feature map as that of $p^{k}_{u}$. The formula is presented below:
\begin{equation} \label{Eq_12}
\begin{aligned}
& (j_{(k,u)}, c_{(k,u)}) = \mathop{\arg\min}\limits_{(j, c) \in \Delta} \left\| M^{c}(x_{j}) - p^{k}_{u} \right\|_{2}, \\
& \Delta = \{(j,c)| y_j = k \land 1 \leq c \leq C_{1} \}, \\
& I(p^{k}_{u}) = B(M^{c_{(k,u)}}(x_{j_{(k,u)}})) \odot x_{j_{(k,u)}}.
\end{aligned}
\end{equation}

Based on Eqs. (\ref{Eq_11}) and (\ref{Eq_12}), any prototype or feature map in ProtoSolo can be visualized and interpreted. To elucidate the decision basis for categorizing the input image $x_{\delta}$ into category $k$ ($k \in \{1,2,...,K\}$), it is essential to identify the key prototype $p^{k}_{u^{\delta}_{k}}$ and locate the feature map $M^{c^{\delta}_{k}}(x_{\delta})$ that exhibits the highest similarity to $p^{k}_{u^{\delta}_{k}}$, as calculated below:
\begin{equation} \label{Eq_13}
\begin{aligned}
u^{\delta}_{k} & = \mathop{\arg\max}\limits_{1 \leq u \leq U}g_{p^{k}_{u}}(x_{\delta}), \\
c^{\delta}_{k} & = \mathop{\arg\min}\limits_{1 \leq c \leq C_{1}}||M^{c}(x_{\delta}) - p^{k}_{u^{\delta}_{k}}||_{2}.
\end{aligned}
\end{equation}

ProtoSolo acquires $x_{\delta}$, $I(M^{c^{\delta}_{k}}(x_{\delta}))$, $I(p^{k}_{u^{\delta}_{k}})$, $G_{k}(x_{\delta})$, $w^{(k,k)}$, and $P(y^{k}|x_{\delta})$ by indexing $u^{\delta}_{k}$ and $c^{\delta}_{k}$, thereby forming a local interpretation for category $k$. The visual explanations $\{\{I(p^{k}_{u})\}_{u=1}^{U}\}_{k=1}^{K}$ of all prototypes across all categories constitute a global explanation within ProtoSolo.

\section{Experiments} \label{Section_4}

The experiments were performed on the CUB-200-2011 (CUB) \cite{wah2011caltech} and Stanford Cars (Car) \cite{krause20133d} public fine-grained image classification datasets. Following a previous study \cite{wang2021interpretable}, bounding-box-cropped images were employed, and augmentations such as random rotation, skew, shear, and horizontal flips were applied. All images were resized to $224 \times 224$ pixels. ProtoSolo was then compared with both classical and advanced prototype networks, including ProtoPNet, ProtoTree, ProtoPShare, ProtoPool, TesNet, D-ProtoPNet, ST-ProtoPNet, PIP-Net, and ProtoArgNet.

\subsection{Implementation Details} \label{Section_41}

The backbone network $f_{b}$ utilizes a ResNet50 \cite{he2016deep} encoder pretrained on ImageNet \cite{deng2009imagenet}. The shaping network $f_{s}$ contains two $1 \times 1$ convolutional layers, each featuring ReLU activation \cite{huang2017densely}. The input image $x$ has dimensions $H = W = 224$ and $C = 3$. Feature maps $M(x)$ are sized $H_{1}=W_{1}=7$ with $C_{1}=64$. The parameters are set as $\epsilon = 10^{-4}$, $\lambda_{1} = 0.8$, $\lambda_{2} = -0.08$, $\lambda_{3} = 10^{-4}$, and $\kappa = 95\%$. The CUB and Cars datasets contain $K=200$ and $K=196$ categories respectively, with each category having $U=10$ prototypes. ProtoSolo adopts ProtoPNet's training configuration, network initialization, and GPU settings.

\subsection{Evaluation Indicators} \label{Section_42}

Top-1 accuracy \cite{singh2021these} was used to assess classification performance. To validate the removal of the prototype projection operation, we used four metrics to evaluate the similarity between the trained prototype and its closest feature map or feature vector: cosine similarity (COS) \cite{jindal2022joyful}, euclidean distance (ED) \cite{jindal2022joyful}, pearson correlation coefficient (PCC) \cite{jindal2022joyful}, and jaccard similarity (JS) \cite{jindal2022joyful}. The prototype quality was evaluated by calculating the percentage of prototypes where the target foreground area's proportion within the bounding box (i.e., precision (Pr) \cite{singh2021these}) exceeded a predefined threshold. We introduce prototype compactness (PC) to measure the number of prototypes used to explain a single category, thereby assessing the cognitive complexity of the explanation.

\begin{table}[t]
	\centering
	\caption{Prototype compactness (PC) and accuracy (\%) of ProtoSolo and comparison methods on the CUB/Car dataset.}
	\begin{tabular}{@{}lcc|cc@{}}
		\hline
		Dataset & \multicolumn{2}{c}{CUB-200-2011} & \multicolumn{2}{c}{Stanford Cars} \\
		\hline
		Method & PC & Accuracy & PC & Accuracy \\
		\hline 
		ProtoPNet & 10 & 79.2$\pm$0.1 & 10 & 86.1$\pm$0.1 \\
		TesNet & 10 & 83.0$\pm$0.2 & 10 & 88.5$\pm$0.2 \\
		D-ProtoPNet & 54 & 83.4$\pm$0.1 & 54 & 88.6$\pm$0.2 \\
		ST-ProtoPNet & 10 & 83.6$\pm$0.2 & 10 & 88.7$\pm$0.2 \\
		ProtoTree & 8.3 & 82.2$\pm$0.7 & 8.5 & 86.6$\pm$0.2  \\
		ProtoPShare & 400 & 74.7$\pm$0.2 & 480 & 86.4$\pm$0.2  \\
		ProtoPool & 202 & 80.3$\pm$0.2 & 195 & 88.9$\pm$0.1  \\
		PIP-Net & 12 & 82.0$\pm$0.3 & 11 & 86.5$\pm$0.3  \\
		ProtoArgNet & 49 & \textbf{85.4$\pm$0.2} & 49 & 89.3$\pm$0.3  \\
		ProtoSolo (ours) & \textbf{1} & 74.9$\pm$0.1 & \textbf{1} & \textbf{89.5$\pm$0.2}  \\
		\hline
	\end{tabular}
	\label{tab:1}
\end{table}

\begin{figure*}[!t]
	\centering
	{\includegraphics[width=1.0\linewidth]{{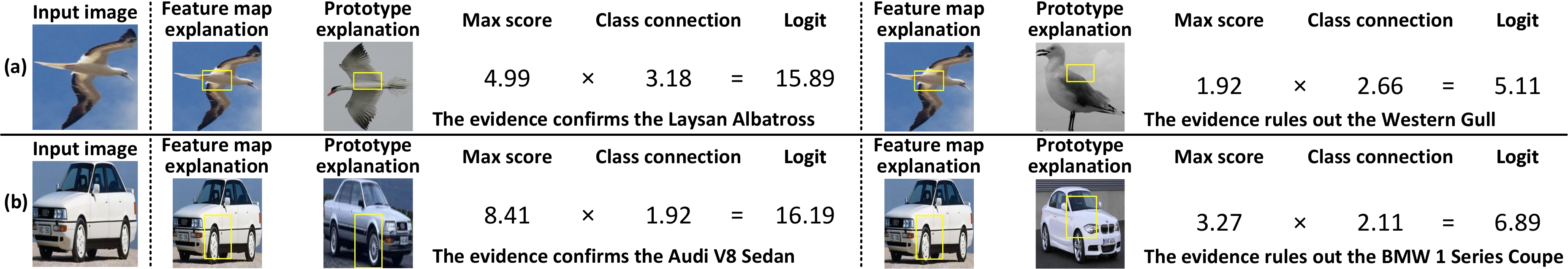}}%
	}
	\caption{Examples of local classification explanations by ProtoSolo for Laysan Albatross and Audi V8 Sedan images.}
	\label{fig_4}
\end{figure*}

\subsection{Recognition Results} \label{Section_43}

Table \ref{tab:1} lists the classification accuracy and the required number of prototypes (PC) for single-class explanations using ProtoSolo and existing methods on the CUB and Cars datasets \cite{ayoobi2025protoargnet}. On the CUB dataset, ProtoSolo achieved 74.9\% accuracy; on Cars, its accuracy outperformed all compared methods. Existing prototype networks require eight or more prototypes to explain single-class decisions. In contrast, ProtoSolo uses a single prototype to explain its categorization of a test image, thereby minimizing cognitive complexity for explanation. As explaining a single class demands at least one unique prototype, ProtoSolo reaches theoretical optimality in cognitive complexity. This demonstrates that, while retaining competitive classification performance, our approach significantly reduces the cognitive complexity of model interpretation, rendering the decision rationale more understandable.

\subsection{Reasoning Process} \label{Section_44}

Figure \ref{fig_4} illustrates the interpretable reasoning process of ProtoSolo for bird and car images, emphasizing its low cognitive complexity. The model classifies by calculating the maximum similarity between the input image $x$'s latent features $\{M^{c}(x)\}_{c=1}^{C_{1}}$ and the prototype set $\{p^{k}_{u}\}_{u=1}^{U}$ of category $k$. Figure \ref{fig_4}(a) demonstrates that the input image's bird body features achieved a maximum score of 4.99 against the Laysan Albatross prototype. After passing through the fully connected layer, the corresponding logit was 15.89. Concurrently, the model determined that the similarity between the input image and the Western Gull's birdback prototype was low at 1.92, resulting in a logit of only 5.11. In summary, ProtoSolo generated a high-confidence prediction based on the strong similarity between the input features and the Laysan Albatross prototype, while the low similarity with the Western Gull prototype led to a lower-confidence prediction, thereby correctly classifying the image as a Laysan Albatross. Its core innovation is that each category's decision relies solely on a single-prototype activation, minimizing cognitive complexity. Figure \ref{fig_4}(b) further illustrates that ProtoSolo can identify a single key prototype to distinguish between categories, ensuring that the reasoning process is explainable and imposes a low cognitive burden. The analysis of the trained fully connected layer weight set $\{\{w^{(t,k)}\}_{k=1}^{K}\}_{t=1}^{K}$ reveals that it satisfies the following properties: $\mathop{\max}\limits_{1 \leq t \leq K} \mathop{\max}\limits_{k \neq t} |w^{(t,k)}| < 5 \times 10^{-3}$, $\mathop{\min}\limits_{1 \leq t \leq K} w^{(t,t)} > 1$, and $\sum_{t=1}^{K}\sum_{k \neq t}|w^{(t,k)}| \approx 0$. This is consistent with the training objective of the loss $L_{w}$ in equation (\ref{Eq_8}), where the model prediction $P(y^{t}|x)$ for category $t$ depends solely on the maximum score $G_{t}(x)$: $P(y^{t}|x) = \sum_{k=1}^{K}w^{(t,k)}G_{k}(x) = \sum_{k\neq t}w^{(t,k)}G_{k}(x) + w^{(t,t)}G_{t}(x) \approx w^{(t,t)}G_{t}(x)$. ProtoSolo's inherent interpretability is consistent with established prototype network theory.

\begin{figure}[!t]
	\centerline{\includegraphics[width=\columnwidth]{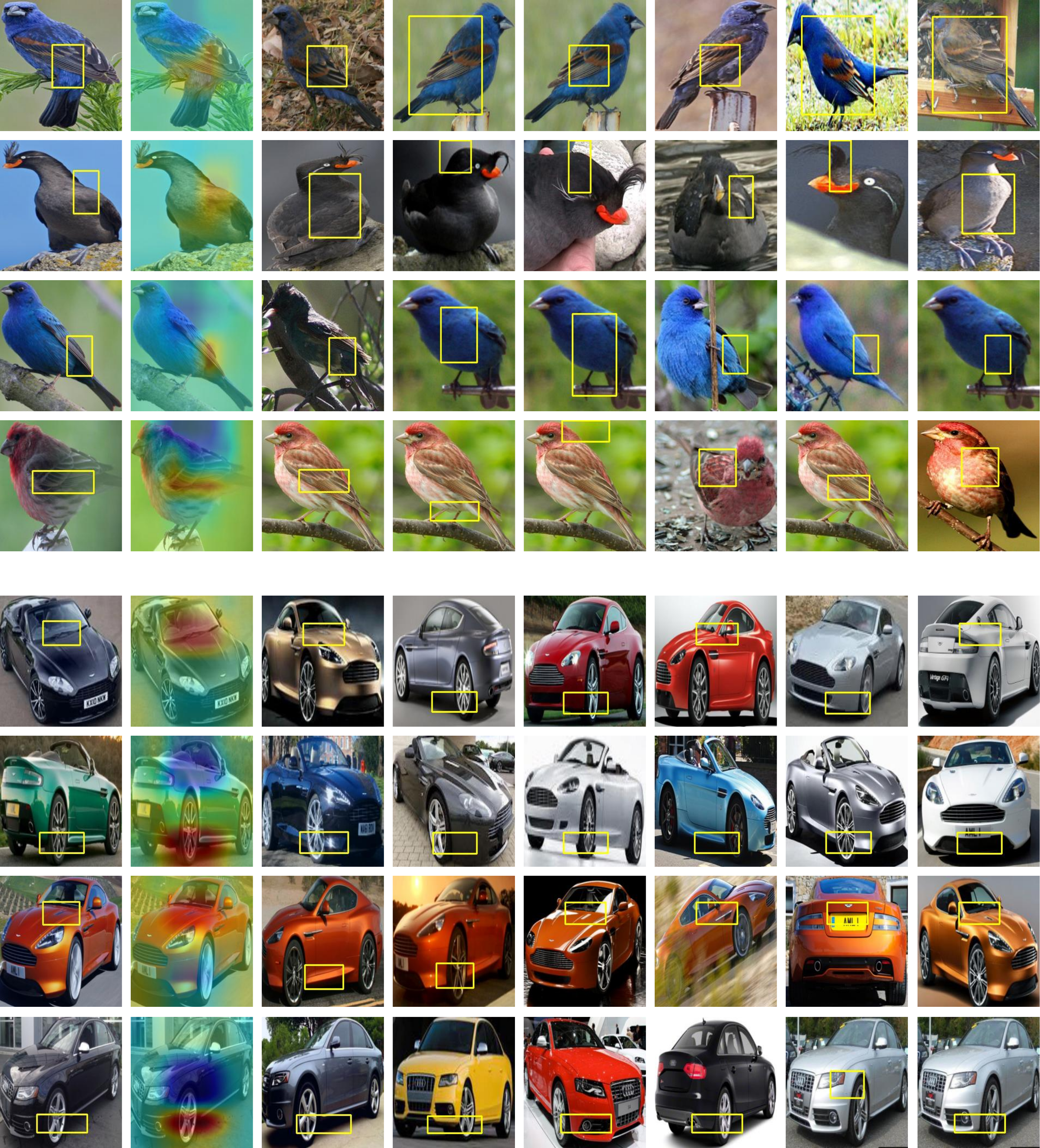}}
	\caption{Example of ProtoSolo visualization explanation: (first column) explanation of the decision region; (second column) feature map visualization; (third to eighth columns) explanation of top-6 in-class prototypes.}
	\label{fig_5}
\end{figure}

Figure \ref{fig_5} displays prototypes across various categories alongside their respective classification feature maps, highlighting the effectiveness of ProtoSolo in interpreting feature maps and prototypes. The first column shows the input image and its classification evidence, with yellow boxes emphasizing critical regions that influence the decision. The second column illustrates the fusion of the input image and the decision-related feature map, where highly activated regions indicate features involved in classification. Columns three to eight feature the representative prototypes derived from the training set for category determination. These results confirm that the yellow boxes successfully identify features supporting classification, thereby validating the effectiveness of the learned prototypes. As the CNN-generated feature map reflects responses to specific pattern mappings within the input image, using it for similarity comparison and prototype learning is appropriate for ProtoSolo. In summary, ProtoSolo necessitates only a single prototype and its corresponding feature map for local explanations and can utilize the entire set of category prototypes for global explanations. This method provides local and global explanation capabilities comparable to existing prototype networks while maintaining minimal cognitive complexity.

\subsection{Non-projection Prototype Learning Testing} \label{Section_45}

To validate our approach, we excluded the prototype projection during the training phase. Despite this exclusion, the model prototype sustained performance similar to the projected prototype and remained aligned with image features. Our experimental setup is as follows.

Initially, we trained the ProtoPNet and ProtoSolo for ten epochs without the projection step, optimizing only the fully connected layer until the losses $L_{w}$ and $L_{total}$ converged. Subsequently, we calculated the similarity between each prototype and its most similar target feature (either a feature map or a full-channel feature vector to be projected) in the training set using the COS, ED, PCC, and JS metrics. The results on the left side of Table \ref{tab:2} show that prototypes without projection closely matched their target features. This indicates that even without the projection step, prototypes can effectively associate with image features through training loss $L_{total}$ and maintain their visual interpretability.

\begin{table}[t]
	\centering
	\caption{Similarity and classification accuracy (\%) (P: projection, NP: non-projection).}
	\resizebox{1.0\columnwidth}{!}{
	\begin{tabular}{@{}lcccc|cc@{}}
		\hline
		Evaluation & COS & ED & PCC & JS & \multicolumn{2}{c}{Accuracy} \\
		\hline
		Method & NP & NP & NP & NP & P & NP\\
		\hline
		Dataset: CUB\\  
		ProtoPNet & 0.93 & 1.36 & 0.75 & 0.77 & 79.2$\pm$0.1 & 79.8$\pm$0.1\\
		ProtoSolo & 0.99 & 0.01 & 0.99 & 0.99 & 74.8$\pm$0.1 & 74.9$\pm$0.1\\
		\hline
		Dataset: Car\\  
		ProtoPNet & 0.95 & 1.16 & 0.82 & 0.81 & 86.1$\pm$0.1 & 89.9$\pm$0.1 \\
		ProtoSolo & 0.99 & 0.01 & 0.99 & 0.99 & 89.1$\pm$0.1 & 89.5$\pm$0.2\\
		\hline
	\end{tabular}}
	\label{tab:2}
\end{table}

To evaluate the impact of the prototype projection on classification performance, we compared the accuracy of ProtoPNet and ProtoSolo with projection (P) and non-projection (NP) during training, as illustrated on the right side of Table \ref{tab:2}. The results demonstrated that network classification accuracy improved when the prototype projection was omitted, suggesting that avoiding network structure modifications due to projections can enhance performance. Therefore, our proposed method of omitting the projection operation increases the accuracy of the prototype network while preserving its interpretability.

\begin{table}[t]
	\centering
	\caption{Percentage of ProtoSolo prototypes exceeding precision (Pr) thresholds (P: projection, NP: non-projection).}
	\resizebox{1.0\columnwidth}{!}{
		\begin{tabular}{@{}lccccc@{}}
			\hline
			Evaluation & \multicolumn{5}{c}{Percentage} \\
			\hline
			Method & Pr$>$10\% & Pr$>$20\% & Pr$>$30\% & Pr$>$40\% & Pr$>$50\% \\
			\hline
			ProtoSolo (P) & 84\% & 78\% & 71\% & 67\% & 61\% \\
			ProtoSolo (NP) & 89\% & 85\% & 80\% & 77\% & 70\% \\
			\hline
		\end{tabular}}
		\label{tab:3}
	\end{table}

Additionally, to assess the quality of the prototype patches in visual explanations with or without prototype projections, we introduced the precision (Pr) metric, which measures the proportion of the ground-truth foreground segmentation area of the image covered by the prototype visualization bounding box. Table \ref{tab:3} displays the percentages of prototype Pr values exceeding the predefined thresholds (10\%, 20\%, 30\%, 40\%, and 50\%) for the ProtoSolo model trained on CUB-200-2011, utilizing projection (P) and non-projection (NP) prototypes. With prototype projection, at least 84\% of the prototypes covered over 10\% of the foreground area, and 61\% achieved 50\% coverage, indicating that their placement areas generally provided a discriminative classification basis. Conversely, the visualization bounding box without prototype projection exhibited greater foreground area coverage. Comprehensive experiments revealed that ProtoSolo without prototype projection performs comparably or better in terms of interpretability, confirming its effectiveness. This approach enhances the classification performance of the method, which effectively prevents performance degradation due to sharp changes in the network weight structure while maintaining strong interpretability.

\subsection{Ablation Study} \label{Section_46}

\begin{table}[t]
	\centering
	\caption{Ablation study results of ProtoPNet/ProtoSolo under single-prototype activation (SA), feature-map-based comparison (FMC), and non-projection (NP) prototype learning. $U$ is the number of prototypes per category.}
	\resizebox{1.0\columnwidth}{!}{
	\begin{tabular}{@{}lccccc@{}}
		\hline
		Evaluation & SA & FMC & NP & \multicolumn{2}{c}{Accuracy} \\
		\hline
		Method & & & & \multicolumn{1}{c}{CUB} & \multicolumn{1}{c}{Cars} \\
		\hline
		ProtoPNet ($U=1$) & & & & 72.6$\pm$0.1 & 87.7$\pm$0.1 \\
		ProtoPNet ($U=10$) & $\checkmark$ & & & 72.4$\pm$0.1 & 87.4$\pm$0.1 \\
		ProtoSolo ($U=10$) & $\checkmark$ & $\checkmark$ & & 74.8$\pm$0.1 & 89.1$\pm$0.1 \\
		ProtoSolo ($U=10$) & $\checkmark$ & $\checkmark$ & $\checkmark$ & \textbf{74.9$\pm$0.1} & \textbf{89.5$\pm$0.2} \\
		\hline
	\end{tabular}}
	\label{tab:4}
\end{table}

To evaluate the effectiveness of single-prototype activation (SA), feature-map-based comparison (FMC), and non-projection (NP) prototype learning operations, Table \ref{tab:4} presents four sets of model results: the first row represents ProtoPNet with a single prototype per class; the second row corresponds to ProtoPNet with SA, employing ten prototypes per class; the third row depicts ProtoSolo with the projection (P) prototype learning; and the fourth row illustrates ProtoSolo with the non-projection (NP) prototype learning.

Rows 1 and 2 in Table \ref{tab:4} demonstrate that, although a single prototype ($U=1$) per class in ProtoPNet facilitates local explanations, its global interpretability is limited due to restricted diversity of prototypes for each class. The introduction of SA ($U=10$) enhances prototype diversity within each class while preserving classification performance, thereby increasing category comprehensibility. Comparing rows 2 and 3 in Table \ref{tab:4}, ProtoSolo which utilizes the feature map for similarity instead of the full-channel feature vector enhances classification accuracy by 2.4\% and 1.7\% over ProtoPNet, respectively. This demonstrates that feature maps with richer information provide a more robust foundation for classifying single-prototype activation networks. Rows 3 and 4 in Table \ref{tab:4} show ProtoSolo with non-projection prototype learning improves classification accuracy. This method prevents performance degradation and reduces computational overhead by avoiding abrupt changes in the network structure caused by projection operations.

\section{Conclusion} \label{Section_5}

This study introduces ProtoSolo, which is a prototype-based interpretable neural network that makes decisions solely based on the similarity score between a single prototype and an input image. This approach simplifies the complexity of traditional prototype networks in explaining individual categories. ProtoSolo employs a feature map instead of a full-channel feature vector for similarity calculation and prototype learning, offering a richer informational base for single-prototype activation networks and presenting a novel perspective on decision interpretation. Additionally, the non-projection prototype learning method effectively preserves model interpretability while preventing classification performance decline caused by sudden network structure changes. ProtoSolo provides a concise, efficient, comprehensible, and extensible interpretable framework, thereby paving the way for the development of future interpretable models.


\bibliographystyle{plain} 

\bibliography{DProtoNet} 

\begin{thebibliography}{10}

\bibitem{ayoobi2025protoargnet}
Hamed Ayoobi, Nico Potyka, and Francesca Toni.
\newblock Protoargnet: Interpretable image classification with super-prototypes
  and argumentation.
\newblock In {\em Proceedings of the AAAI Conference on Artificial
  Intelligence}, volume~39, pages 1791--1799, 2025.

\bibitem{chen2019looks}
Chaofan Chen, Oscar Li, Daniel Tao, Alina Barnett, Cynthia Rudin, and
  Jonathan~K Su.
\newblock This looks like that: deep learning for interpretable image
  recognition.
\newblock {\em Advances in neural information processing systems}, 32, 2019.

\bibitem{deng2009imagenet}
Jia Deng, Wei Dong, Richard Socher, Li-Jia Li, Kai Li, and Li~Fei-Fei.
\newblock Imagenet: A large-scale hierarchical image database.
\newblock In {\em 2009 IEEE conference on computer vision and pattern
  recognition}, pages 248--255. IEEE, 2009.

\bibitem{di2025ante}
Antonio Di~Marino, Vincenzo Bevilacqua, Angelo Ciaramella, Ivanoe De~Falco, and
  Giovanna Sannino.
\newblock Ante-hoc methods for interpretable deep models: A survey.
\newblock {\em ACM Computing Surveys}, 57(10):1--36, 2025.

\bibitem{donnelly2022deformable}
Jon Donnelly, Alina~Jade Barnett, and Chaofan Chen.
\newblock Deformable protopnet: An interpretable image classifier using
  deformable prototypes.
\newblock In {\em Proceedings of the IEEE/CVF Conference on Computer Vision and
  Pattern Recognition}, pages 10265--10275, 2022.

\bibitem{he2016deep}
Kaiming He, Xiangyu Zhang, Shaoqing Ren, and Jian Sun.
\newblock Deep residual learning for image recognition.
\newblock In {\em Proceedings of the IEEE conference on computer vision and
  pattern recognition}, pages 770--778, 2016.

\bibitem{hearst1998support}
Marti~A. Hearst, Susan~T Dumais, Edgar Osuna, John Platt, and Bernhard
  Scholkopf.
\newblock Support vector machines.
\newblock {\em IEEE Intelligent Systems and their applications}, 13(4):18--28,
  1998.

\bibitem{huang2017densely}
Gao Huang, Zhuang Liu, Laurens Van Der~Maaten, and Kilian~Q Weinberger.
\newblock Densely connected convolutional networks.
\newblock In {\em Proceedings of the IEEE conference on computer vision and
  pattern recognition}, pages 4700--4708, 2017.

\bibitem{jindal2022joyful}
Anisha Jindal, Naveen Sharma, and Vijay Verma.
\newblock Joyful jaccard: An analysis of jaccard-based similarity measures in
  collaborative recommendations.
\newblock In {\em International Conference on Artificial Intelligence and
  Sustainable Engineering: Select Proceedings of AISE 2020, Volume 1}, pages
  29--41. Springer, 2022.

\bibitem{keswani2022proto2proto}
Monish Keswani, Sriranjani Ramakrishnan, Nishant Reddy, and Vineeth~N
  Balasubramanian.
\newblock Proto2proto: Can you recognize the car, the way i do?
\newblock In {\em Proceedings of the IEEE/CVF Conference on Computer Vision and
  Pattern Recognition}, pages 10233--10243, 2022.

\bibitem{krause20133d}
Jonathan Krause, Michael Stark, Jia Deng, and Li~Fei-Fei.
\newblock 3d object representations for fine-grained categorization.
\newblock In {\em Proceedings of the IEEE international conference on computer
  vision workshops}, pages 554--561, 2013.

\bibitem{li2023g}
Xuhong Li, Haoyi Xiong, Xingjian Li, Xiao Zhang, Ji~Liu, Haiyan Jiang, Zeyu
  Chen, and Dejing Dou.
\newblock G-lime: Statistical learning for local interpretations of deep neural
  networks using global priors.
\newblock {\em Artificial Intelligence}, 314:103823, 2023.

\bibitem{liu2024unbiased}
Yajing Liu, Shijun Zhou, Xiyao Liu, Chunhui Hao, Baojie Fan, and Jiandong Tian.
\newblock Unbiased faster r-cnn for single-source domain generalized object
  detection.
\newblock In {\em Proceedings of the IEEE/CVF Conference on Computer Vision and
  Pattern Recognition}, pages 28838--28847, 2024.

\bibitem{nauta2023pip}
Meike Nauta, J{\"o}rg Schl{\"o}tterer, Maurice van Keulen, and Christin
  Seifert.
\newblock Pip-net: Patch-based intuitive prototypes for interpretable image
  classification.
\newblock In {\em Proceedings of the IEEE/CVF Conference on Computer Vision and
  Pattern Recognition}, pages 2744--2753, 2023.

\bibitem{nauta2021neural}
Meike Nauta, Ron Van~Bree, and Christin Seifert.
\newblock Neural prototype trees for interpretable fine-grained image
  recognition.
\newblock In {\em Proceedings of the IEEE/CVF Conference on Computer Vision and
  Pattern Recognition}, pages 14933--14943, 2021.

\bibitem{peng2024decoupling}
Yitao Peng, Lianghua He, Die Hu, Yihang Liu, Longzhen Yang, and Shaohua Shang.
\newblock Decoupling deep learning for enhanced image recognition
  interpretability.
\newblock {\em ACM Transactions on Multimedia Computing, Communications and
  Applications}, 20(10):1--24, 2024.

\bibitem{petsiuk2021black}
Vitali Petsiuk, Rajiv Jain, Varun Manjunatha, Vlad~I Morariu, Ashutosh Mehra,
  Vicente Ordonez, and Kate Saenko.
\newblock Black-box explanation of object detectors via saliency maps.
\newblock In {\em Proceedings of the IEEE/CVF Conference on Computer Vision and
  Pattern Recognition}, pages 11443--11452, 2021.

\bibitem{rymarczyk2022interpretable}
Dawid Rymarczyk, {\L}ukasz Struski, Micha{\l} G{\'o}rszczak, Koryna
  Lewandowska, Jacek Tabor, and Bartosz Zieli{\'n}ski.
\newblock Interpretable image classification with differentiable prototypes
  assignment.
\newblock In {\em Computer Vision--ECCV 2022: 17th European Conference, Tel
  Aviv, Israel, October 23--27, 2022, Proceedings, Part XII}, pages 351--368.
  Springer, 2022.

\bibitem{rymarczyk2021protopshare}
Dawid Rymarczyk, {\L}ukasz Struski, Jacek Tabor, and Bartosz Zieli{\'n}ski.
\newblock Protopshare: Prototypical parts sharing for similarity discovery in
  interpretable image classification.
\newblock In {\em Proceedings of the 27th ACM SIGKDD Conference on Knowledge
  Discovery \& Data Mining}, pages 1420--1430, 2021.

\bibitem{shrikumar2017learning}
Avanti Shrikumar, Peyton Greenside, and Anshul Kundaje.
\newblock Learning important features through propagating activation
  differences.
\newblock In {\em International conference on machine learning}, pages
  3145--3153. PMLR, 2017.

\bibitem{singh2021these}
Gurmail Singh and Kin-Choong Yow.
\newblock These do not look like those: An interpretable deep learning model
  for image recognition.
\newblock {\em IEEE Access}, 9:41482--41493, 2021.

\bibitem{sundararajan2017axiomatic}
Mukund Sundararajan, Ankur Taly, and Qiqi Yan.
\newblock Axiomatic attribution for deep networks.
\newblock In {\em International conference on machine learning}, pages
  3319--3328. PMLR, 2017.

\bibitem{wah2011caltech}
Catherine Wah, Steve Branson, Peter Welinder, Pietro Perona, and Serge
  Belongie.
\newblock The caltech-ucsd birds-200-2011 dataset.
\newblock 2011.

\bibitem{wang2023learning}
Chong Wang, Yuyuan Liu, Yuanhong Chen, Fengbei Liu, Yu~Tian, Davis McCarthy,
  Helen Frazer, and Gustavo Carneiro.
\newblock Learning support and trivial prototypes for interpretable image
  classification.
\newblock In {\em Proceedings of the IEEE/CVF International Conference on
  Computer Vision}, pages 2062--2072, 2023.

\bibitem{wang2021interpretable}
Jiaqi Wang, Huafeng Liu, Xinyue Wang, and Liping Jing.
\newblock Interpretable image recognition by constructing transparent embedding
  space.
\newblock In {\em Proceedings of the IEEE/CVF International Conference on
  Computer Vision}, pages 895--904, 2021.

\bibitem{xu2024sctnet}
Zhengze Xu, Dongyue Wu, Changqian Yu, Xiangxiang Chu, Nong Sang, and Changxin
  Gao.
\newblock Sctnet: Single-branch cnn with transformer semantic information for
  real-time segmentation.
\newblock In {\em Proceedings of the AAAI conference on artificial
  intelligence}, volume~38, pages 6378--6386, 2024.

\bibitem{zheng2022shap}
Quan Zheng, Ziwei Wang, Jie Zhou, and Jiwen Lu.
\newblock Shap-cam: Visual explanations for convolutional neural networks based
  on shapley value.
\newblock In {\em European conference on computer vision}, pages 459--474.
  Springer, 2022.

\end{thebibliography}

\vfill


\end{document}